\documentclass{article} 
\usepackage[preprint]{colm2025_conference}

\usepackage{microtype}
\usepackage{hyperref}
\usepackage{url}
\usepackage{booktabs}
\usepackage{graphicx}
\usepackage{subcaption}
\usepackage{amsmath}
\usepackage{multirow}

\usepackage{lineno}

\definecolor{darkblue}{rgb}{0, 0, 0.5}
\hypersetup{colorlinks=true, citecolor=darkblue, linkcolor=darkblue, urlcolor=darkblue}

\title{Faster MoE LLM Inference for Extremely Large Models}

 \author{Haoqi Yang$^{\dagger}$, Luohe Shi$^{\dagger}$, Qiwei Li, Zuchao Li\thanks{Corresponding Author, $^{\dagger}$Equal Contribution},\ \  Ping Wang, Bo Du\\
Wuhan University \\
Wuhan, 430072, P. R. China \\
\And
Mengjia Shen \\
Wuhan Second Ship Design And Research Institute \\
Wuhan, 430010, P. R. China \\
\AND
Hai Zhao \\
Shanghai Jiao Tong University \\
Shanghai, 200240, P. R. China \\
}

\begin{document}

\ifcolmsubmission
\linenumbers
\fi

\maketitle

\begin{abstract}
Sparse Mixture of Experts (MoE) large language models (LLMs) are gradually becoming the mainstream approach for ultra-large-scale models.
Existing optimization efforts for MoE models have focused primarily on coarse-grained MoE architectures. With the emergence of DeepSeek Models, fine-grained MoE models are gaining popularity, yet research on them remains limited. 
Therefore, we want to discuss the efficiency dynamic under different service loads. 
Additionally, fine-grained models allow deployers to reduce the number of routed experts, both activated counts and total counts, raising the question of how this reduction affects the trade-off between MoE efficiency and performance.
Our findings indicate that while deploying MoE models presents greater challenges, it also offers significant optimization opportunities. Reducing the number of activated experts can lead to substantial efficiency improvements in certain scenarios, with only minor performance degradation. Reducing the total number of experts provides limited efficiency gains but results in severe performance degradation. Our method can increase throughput by at least 10\% without any performance degradation. Overall, we conclude that MoE inference optimization remains an area with substantial potential for exploration and improvement.

\end{abstract}

\section{Introduction}

The introduction of the Transformer~\citep{NIPS2017_3f5ee243} has significantly improved the training efficiency of sequence models, making the training of ultra-large-scale models feasible. Building upon this foundation, GPT-style decoder-only Transformer models~\citep{radford2018improving, radford2019language, NEURIPS2020_1457c0d6, DBLP:journals/corr/abs-2303-08774}, guided by scaling laws~\citep{DBLP:journals/corr/abs-2001-08361}, have rapidly gained prominence~\citep{DBLP:journals/corr/abs-2302-13971, DBLP:journals/corr/abs-2307-09288, DBLP:journals/corr/abs-2407-21783}. Their exceptional scalability has made them the mainstream choice for constructing large-scale generative language models~\citep{DBLP:journals/jmlr/ChowdheryNDBMRBCSGSSTMRBTSPRDHPBAI23}. To further expand model size while controlling both inference and training costs, one promising direction is the sparsification of the feed-forward network (FFN), which constitutes the majority of parameters in LLMs. 

The FFN module in LLMs is typically implemented as either a Multi-Layer Perceptron (MLP, \citealp{MURTAGH1991183}) or a Gated Linear Unit (GLU, \citealp{DBLP:conf/icml/DauphinFAG17}). Both first project the hidden state to an intermediate state via an upsampling transformation, followed by an activation function, and then downsampled to produce the new hidden state.
The intermediate state in MLP and GLU can be easily partitioned: a large one can be divided into multiple smaller ones while maintaining strict equivalence. This enables a strategy where an excessively large intermediate state is split into multiple smaller states, with only a subset of them being selectively activated through gating. 
Parameters corresponding to each partition is referred to as an "expert", and this architecture is known as the Sparse Mixture-of-Experts (MoE, \citealp{DBLP:conf/iclr/ShazeerMMDLHD17}). MoE was demonstrated to be effective in Transformer architectures by \citeauthor{DBLP:journals/jmlr/FedusZS22}. Mixtral-8x7B~\citep{DBLP:journals/corr/abs-2401-04088} is the most famous large-scale open-source MoE LLM.

The deployment of Large Language Models (LLMs) has long been a significant challenge. Due to the autoregressive decoding, model parameters are repeatedly accessed during inference~\citep{280922} but with only a few computation, leading to low arithmetic intensity.
According to the Roofline model~\citep{10.1145/1498765.1498785}, this harms the computational efficiency of the system. 
Several existing works have optimized this aspect by enlarging batch size~\citep{10.1145/3600006.3613165, NEURIPS2024_724be447, ye-etal-2024-chunkattention}. 
While significant progress has been made in improving the training efficiency of MoE models~\citep{DBLP:journals/corr/abs-2103-13262, 10.1145/3503221.3508418, 288691, pmlr-v162-rajbhandari22a}, accelerating MoE modules during inference remains an underexplored area. In particular, the theoretical performance of MoE-based inference systems under varying batch sizes has yet to be thoroughly investigated.

\begin{figure}[t]
    \centering
    \vspace{-1em}
    \includegraphics[width=\linewidth]{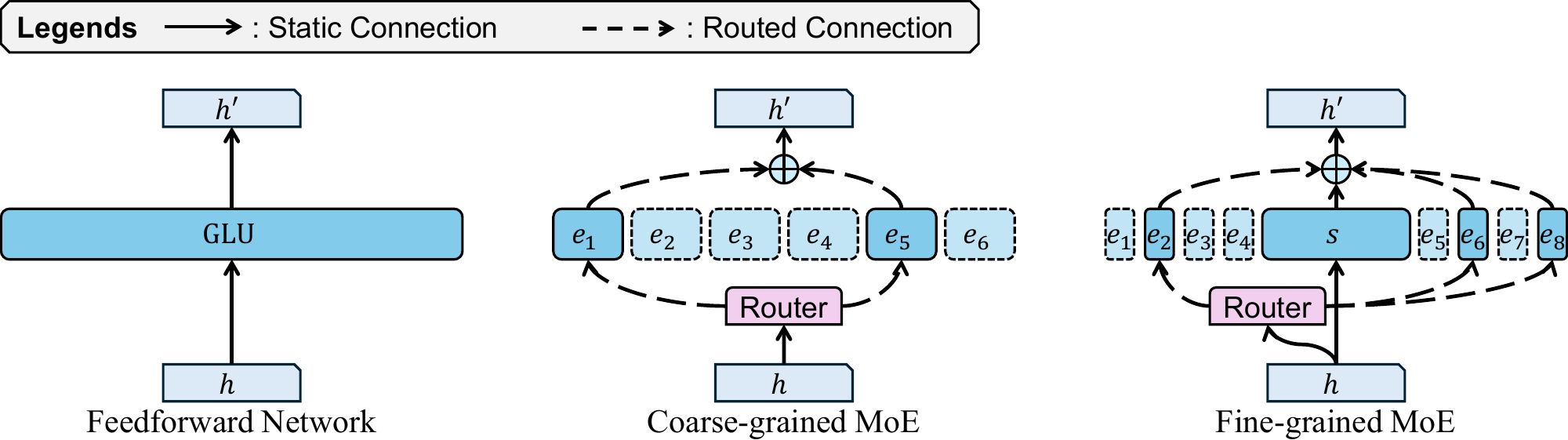}
    \caption{The comparison of FFN, coarse-grained MoE, and fine-grained MoE.}
    \label{fig:comparison}
    \vspace{-1em}
\end{figure}

Traditional MoE models employ a relatively conservative number of experts, typically eight in total and two actived per token. Moreover, these experts are generally initialized from a pre-trained dense small model. \citeauthor{dai-etal-2024-deepseekmoe} introduced several modifications. They significantly increased the number of experts, both total and activated, while reducing the size of each expert. Additionally, all experts were randomly initialized. To mitigate the instability associated with training smaller experts, they introduced a larger shared expert which is always chosen as a backbone. The release of Deepseek V2~\citep{DBLP:journals/corr/abs-2405-04434}, featuring 236 billion parameters, demonstrated the scalability of this paradigm. 
\citeauthor{DBLP:journals/corr/abs-2501-12948} further optimized the expert selection mechanism by incorporating grouping constraints and sigmoid activation, instead of softmax, leading to the development of the Deepseek V3 model. We refer to the deepseek approach as fine-grained MoE, while the conventional MoE paradigm is termed coarse-grained MoE. As shown in Figure~\ref{fig:comparison}.

Fine-grained MoE enhances the model’s adaptability, allowing for more flexible parameter utilization during inference. Studies have explored discarding certain parameters at inference time to improve decoding speed. These approaches include expert pruning, where specific experts are removed and not loaded into memory~\citep{DBLP:journals/corr/abs-2206-00277, DBLP:journals/corr/abs-2409-06211, lu-etal-2024-experts, DBLP:journals/corr/abs-2410-12013, DBLP:journals/corr/abs-2411-01016}, and expert skipping, where the number of activated experts per token is reduced~\citep{lu-etal-2024-experts}. Although some prior work has explored related topics, it has predominantly focused on designs tailored for coarse-grained MoE rather than the more promising fine-grained variant. 

In summary, this paper presents an efficiency analysis of fine-grained MoE models under varying batch sizes during inference. We aim to address the question of whether MoE models can achieve efficiency gains only in large-scale service scenarios. Furthermore, we investigate the impact of different expert reduction strategies on both model performance and computational efficiency to assess whether this direction warrants further exploration, particularly in fine-grained MoE and its sigmoid-activated variant, in which the model exhibits different dynamical behaviors, necessitating a distinct analytical approach.

\section{Background and Related Works}

\subsection{FFN and MoE}

Feed-forward network is usually an MLP or a GLU, given in Equation~\ref{eq:MLPnGLU}, where $\mathrm{ACT}(\cdot)$ denotes the activation function, and $\otimes$ denotes the element-wise product.
\begin{equation}
\label{eq:MLPnGLU}
\begin{aligned}
    \mathrm{MLP}(h) &= W_{\mathrm{d}} \cdot 
        \mathrm{ACT}\left(W_{\mathrm{u}}\cdot h \right) \\
    \mathrm{GLU}(h) &= W_{\mathrm{d}} \cdot \left(
        \mathrm{ACT}\left(W_{\mathrm{u}}\cdot h \right)
        \otimes
        \left( W_{\mathrm{g}} \cdot h \right)
    \right)
\end{aligned}
\end{equation}

We use GLU to represent FFN from now on since it's the more common practice in current days~\citep{DBLP:journals/corr/abs-2002-05202, DBLP:journals/jmlr/ChowdheryNDBMRBCSGSSTMRBTSPRDHPBAI23}.

A mixture of experts layer can be represented as the weighted combination of multiple GLUs, where the weights $r$ are given by a linear classifier called router. We define router logits as $r'$. Moreover, there is usually two function to modify the logits, $F_r(\cdot)$ to manipulate Top$k$ selection for load balancing, and $F_w(\cdot)$ to normalize logits into the final weights. An MoE layer with $n_e$ experts and $n_a$ activated is given in Equation~\ref{eq:MoE}
\begin{equation}
\label{eq:MoE}
\begin{aligned}
    \mathrm{MoE}(h) &= \sum_{i=1}^{n_e}r_i\cdot \mathrm{GLU}_i(h) \\
    r' = W_{\mathrm{r}}\cdot h,\ \ r_i &= \left\{ \begin{aligned}
        F_w(r'_i) &,\ r'_i \in \mathrm{Top}k\left(F_r(r'), k=n_a\right) \\ 
        0 &,\ \mathrm{otherwise}.
    \end{aligned}
    \right.
\end{aligned}
\end{equation}

\subsection{LLM Serving Efficiency}

Although modern LLM service systems offer a variety of system-level objectives (SLOs, \citealp{DBLP:journals/corr/abs-2410-14257}) for performance analysis, we focus on the most fundamental metric, {\bf throughput}, to avoid unnecessary complexity. Throughput reflects the efficiency of computation under fixed input-output sequence lengths, effectively measuring the utilization of the hardware's computational capacity.

Previous works, such as Orca~\citep{280922} and vLLM~\citep{10.1145/3600006.3613165}, have constructed analytical frameworks based on the Roofline~\citep{10.1145/1498765.1498785} model, which suggests that higher overall computational intensity leads to reduced per-token computation time. A core principle in LLM inference optimization is maximizing batch size to enhance computational efficiency. This is exemplified by continuous batching in vLLM and chunk attention~\citep{ye-etal-2024-chunkattention}, which facilitates batch processing of KV cache segments with shared prefix tokens.

\subsection{Model Pruning}

Model pruning refers to the process of reducing a model’s parameter count to enhance inference speed while minimizing performance degradation. Pruning can be performed at different levels of granularity. For instance, LayerSkip~\citep{gromov2025the} enables skipping entire layers, while other methods apply pruning within layers by sparsifying the attention of FFN~\citep{DBLP:conf/icml/FrantarA23, he-etal-2024-chess}. In the context of LLMs, pruning is not limited to model parameters. It can also extend to components such as the KV cache, where various approaches have been proposed to selectively remove stored key-value pairs, further optimizing memory usage and computational efficiency.

For MoE models, related research has primarily focused on expert pruning~\citep{DBLP:journals/corr/abs-2206-00277, DBLP:journals/corr/abs-2411-01016, DBLP:journals/corr/abs-2410-12013, lu-etal-2024-experts, DBLP:journals/corr/abs-2409-06211, DBLP:journals/corr/abs-2404-05089, DBLP:conf/iclr/Li0YS0BC24, chen2025retrainingfree}. For coarse-grained MoE, pruning was a relatively straightforward approach due to the high homogeneity among experts. However, in fine-grained MoE, these methods face significant challenges. Nevertheless, the increased number of experts, both global and active, also presents new opportunities for optimization.

\section{Preliminaries}
\subsection{Notations}

Previously we defined $d$ as the hidden size, $d_i$ as the intermediate size for FFNs, $n_e$ as the expert counts, $n_a$ as the active (chosen) experts per token, $F_r$ as the modifier function of router logits for expert selection, and $F_w$ as the weight modifier. We denote $L$ as the sequence length, results in hidden state $h \in \mathbb{R}^d$, $H \in \mathbb{R}^{d \times L}$, $W_u, W_g \in {R}^{d_i \times d}$, and $ W_d \in {R}^{d \times d_i}$. The memory I/O, FLOPS, and arithmetic intensity (AI) based on the triplet of $(d, d_i, L)$ is given in Equation~\ref{eq:ioflopsai}.
\begin{equation}
\label{eq:ioflopsai}
\begin{aligned}
    \mathrm{I/O}   (d, d_i, L) &= 3d_id + 2L(d + d_i) \\
    \mathrm{FLOPS} (d, d_i, L) &= 6L(d_id) \\
    \mathrm{AI}    (d, d_i, L) &= \frac{6L(d_id)}{3d_id + 2L(d_i + d)}
\end{aligned}
\end{equation}

To better evaluate the size of each state, we further define $d_e$ as the intermediate size of experts, $d_s$ as the intermediate size of of shared expert, and $d_a$ as the activated intermediate size where $d_a = d_e\times n_a$.

\subsection{Evaluation Method}
We selected two representative fine-grained MoE models for evaluation: DeepSeek-V2-Lite and DeepSeek-V3. Their fundamental characteristics are presented in Table~\ref{tab:v2lv3}.

\begin{table}[ht]
    \centering
    \begin{tabular}{l|cc|cccc|c|l}
    \toprule
        Model            & 
            $n_e$ & $n_a$ & $d$    & $d_e$  & $d_s$   & $d_a$   & $d_a/(d_s+d_a)$ &
            $F_w$\\
    \midrule
        Deepseek-V2-Lite & 
            $64$  & $6$   & $2048$ & $1408$ & $10944$ & $8448$  & $45.6\%$ &
            $\mathrm{softmax}$ \\
        Deepseek-V3      & 
            $256$ & $8$   & $7168$ & $2048$ & $18432$ & $16384$ & $47.1\%$ &
            $\mathrm{sigmoid}$ \\
    \bottomrule
    \end{tabular}
    \caption{Model information of Deepseek-V2-Lite and V3.}
    \label{tab:v2lv3}
\end{table}

To evaluate the model's performance, we selected several benchmark datasets, including ARC (Easy and Challenge, \citealp{DBLP:journals/corr/abs-1803-05457}), BoolQ~\citep{clark-etal-2019-boolq}, OpenBookQA (OBQA,~\citealp{mihaylov-etal-2018-suit}), RTE~\citep{DBLP:conf/tac/BentivogliMDDG09}, and Winogrande~\citep{DBLP:journals/cacm/SakaguchiBBC21}. 
If not specified, the performance score is the average of all above benchmarks, with 36 as the baseline (can be achieved through pure guessing).

\subsection{Hardware and Implementation Details}
Implementation details of Section~\ref{sec:3} is given in Appendix~\ref{app:sec3}. Implementation details of Section~\ref{sec:4} and \ref{sec:6} is given in Appendix~\ref{app:sec4}.

It is important to emphasize that one of the most critical controlled variables in our efficiency experiments is the number of input and output tokens, as highlighted in Appendix~\ref{app:iosum}. Unless otherwise specified, all tests were conducted with randomly sampled 1024 input tokens, while the model was instructed to generate an additional 1024 tokens during inference, since increasing the proportion of input tokens can significantly enhance overall throughput.

\section{Severing Efficiency of MoE}
\label{sec:3}

\subsection{Weakened Batch Effect}

\begin{figure}[ht]
    \centering
    \begin{subfigure}{0.32\textwidth}
        \centering
        \caption{Arithmetic intensity}
        \label{fig:AIL}
        \includegraphics[width=\linewidth]{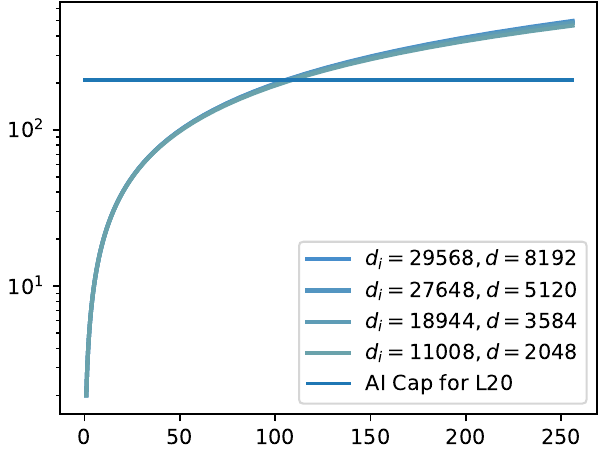}    
    \end{subfigure}
    \begin{subfigure}{0.32\textwidth}
        \centering
        \caption{FFN time per token ($\mu$s)}
        \label{fig:ttp}
        \includegraphics[width=\linewidth]{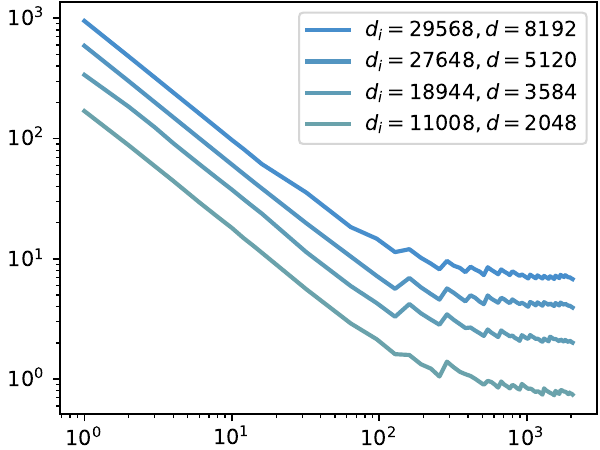}
    \end{subfigure}
    \begin{subfigure}{0.32\textwidth}
        \centering
        \caption{MoE time per token ($\mu$s)}
        \label{fig:33}
        \includegraphics[width=\linewidth]{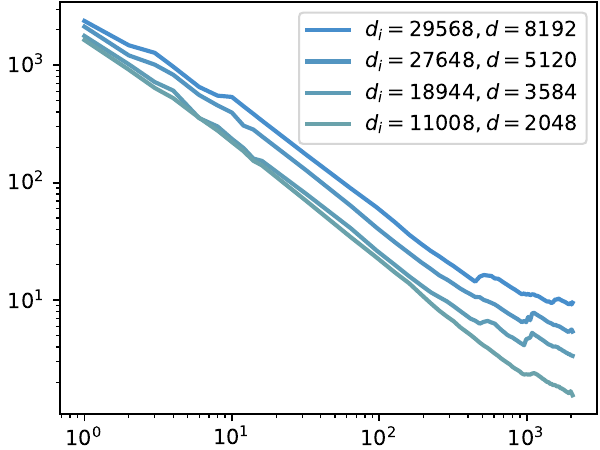}
    \end{subfigure}
    
    
    \caption{Simulation experiment results. X-axis represents sequence length $L$.}
    \label{fig:theory-overall}
\end{figure}

The feed-forward layer constitutes the majority of model parameters, accounting for approximately 66\% in earlier models and up to 88\% in some modern models (\href{https://huggingface.co/Qwen/Qwen2.5-3B}{Qwen-2.5-3B}, \citealp{DBLP:journals/corr/abs-2412-15115}). In a single-batch setting, it dominates computation time. During the prefill phase, all tokens in a sequence pass through the FFN simultaneously, meaning that once parameters are loaded into memory, they are reused multiple times. This amortizes the parameter loading cost across multiple tokens. Figure~\ref{fig:AIL} depicted relationship between $L$ and AI. When the batch size is small, increasing the number of parallel tokens significantly improves system performance. Notably, doubling the number of tokens from one to two incurs virtually no additional latency, despite the computational workload doubling. This phenomenon is illustrated in Figure~\ref{fig:ttp}, showcasing the relationship between $L$ and per-token latency ($\mu$s). It can be observed that at around $L=150$, the efficiency maxed out.

In MoE models, although sparse activation reduces computational demands, resulting in a much lower FLOP requirement than the total parameter count, additional experts must be loaded into memory as the number of tokens increases. Since tokens rarely reuse the same expert within a batch, this creates an additional memory access overhead. Consequently, even without considering scheduling overhead (which can be significant), MoE is inherently not faster than FFN when operating under the same activated parameter count. 

We evaluated the efficiency of the MoE module across different sequence lengths, as shown in Figure~\ref{fig:33}. Our findings indicate that the MoE module incurs higher latency and reaches its peak efficiency more slowly compared to Figure~\ref{fig:ttp}. However, we observed that larger models tend to achieve maximum efficiency more easily, primarily due to their inherently higher AI upper bound.

\subsection{Expert Parallel}

Although MoE is less efficient than FFN in terms of batch processing speedup, it requires significantly less inter-device communication under the same computational demands. When deploying models across multiple computing devices (e.g., GPUs or compute nodes), the most effective method for maximizing hardware utilization is tensor parallelism (TP).

For a dense FFN, TP is implemented by partitioning the intermediate states across multiple devices. Specifically, in a system with $n_d$ devices, each device stores only a fraction of the parameters from $W_\mathrm{\{u,d,g\}}$, corresponding to a subpartition of length $d_i/n_d$ along the intermediate dimension. As previously discussed, the FFN structure allows for straightforward partitioning, enabling computations to be executed in parallel across the $n_d$ devices. However, the primary drawback of TP lies in its high communication overhead. The total inter-device data transfer volume, even in the most optimized scenarios, cannot be lower than $2(n_d-1)Ld$. This is because each hidden state value must be exchanged across all devices twice, once for mapping and once for reducing, significantly limited it's applicability: TP is only seen across GPUs within a node, but not across nodes.

MoE inherently overcomes this communication bottleneck through expert parallelism (EP). In an EP setting, different experts are distributed across different devices, meaning that each token's state is transmitted only to its selected experts. Consequently, in the worst-case scenario, EP requires only $2n_aLd$ communication operations. Given that the typical value of $n_d$ is $8$, whereas $n_a$ is typically $2$ for coarse-grained models, EP reduces data transfer volume to approximately 28\% of that in TP. In fine-grained MoE, a similar optimization can be achieved by grouping experts according to the EP paradigm, where all experts residing on the same device belong to the same group. By constraining each token to select experts from a limited number of groups (e.g., selecting from only $2$ out of $8$ possible groups), we can achieve the same benefits while further optimizing communication overhead.

In a typical multi-GPU setup, intra-node bandwidth using NVLink is approximately $160$ GB/s, whereas inter-node connectivity via InfiniBand achieves around $50$ GB/s, roughly 31\% of intra-node bandwidth~\cite{DBLP:journals/corr/abs-2412-19437}. This indicates that, under such configurations, EP can be effectively implemented across nodes, rather than requiring intra-node TP, while maintaining comparable latency. This property effectively compensates for the efficiency limitations discussed in the previous section: experts within a layer can be distributed across multiple nodes, allowing each node to handle a larger and more concentrated set of requests, thereby increasing arithmetic intensity.








\section{Inference Time Expert Skipping}
\label{sec:4}

Compared to coarse-grained MoE, which typically activates only two experts, fine-grained MoE selects 6 to 8 experts from a pool of 64 to 256. This provides an opportunity to reduce the number of activated experts $n_a$, potentially improving efficiency. However, merely reducing $n_a$ has limited impact on the overall model size, meaning it does not significantly lower the deployment barrier. Additionally, due to the presence of a large shared expert backbone, the actual reduction in computational cost is also relatively constrained.  

Despite these challenges, we aim to explore this aspect from a language modeling perspective. More fine-grained control over $n_a$ can provide insights into the adaptability of MoE models by analyzing their capacity requirements across different components. We conduct our experiments on DeepSeek-V2-Lite and DeepSeek-V3, detailed result can be found in Appendix~\ref{app:dr1}.

\subsection{Efficiency}

\begin{figure}[ht]
    \centering
    \begin{subfigure}{0.48\textwidth}
        \centering
        \caption{Deepseek-V2-Lite}
        \label{fig:v2ltopk}
        \includegraphics[width=\linewidth]{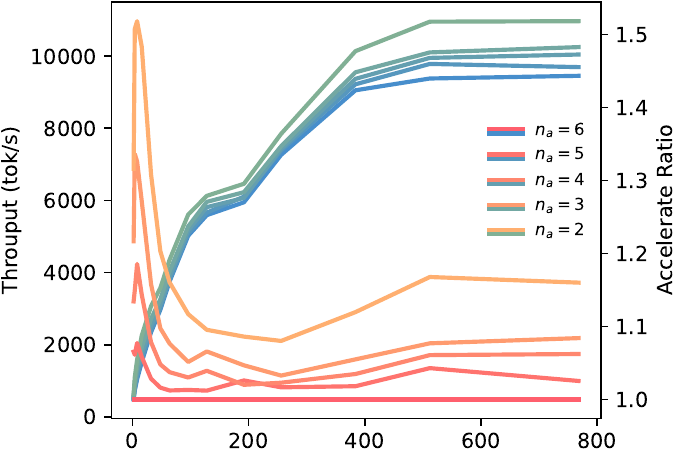}    
    \end{subfigure}
    \hfill
    \begin{subfigure}{0.48\textwidth}
        \centering
        \caption{Deepseek-V3}
        \label{fig:v3topk}
        \includegraphics[width=\linewidth]{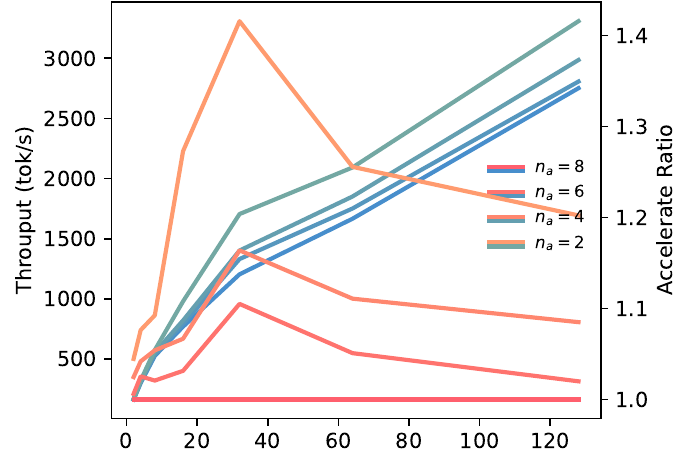}
    \end{subfigure}    
    \caption{How expert skipping influence throughput. X-axis represents the concurrency.}
    \label{fig:efftopk}
\end{figure}

In terms of efficiency, we investigated the impact of varying the number of activated experts ($n_a$) while retaining all experts. To simplify the testing scope and procedure, we used a consistent $n_a$ across all layers, ranging from $2$ to the model’s original $n_a$. Based on this setup, we further evaluated the models under different request loads to analyze their performance. Our experiment results are depicted in Figure~\ref{fig:efftopk}.

By analyzing the throughput trend, we observed that reducing the number of activated experts $n_a$ has no significant impact on the number of concurrent requests required to reach peak throughput. However, the speedup ratio curve presents a much more complex pattern.

Notably, we observe substantial acceleration at both low and high concurrency levels, while the acceleration effect is limited at moderate concurrency. In fact, at low concurrency, the speedup ratio can reach 50\% (when $n_a=2$), surpassing the theoretical compute reduction upper bound for fine-grained MoE without considering MLA layers ($d_a/(d_s+d_a)$, 45\%). This is because at low concurrency, the system is memory I/O-bound, and reducing $n_a$ immediately lowers the required parameter loading, leading to a higher proportion of acceleration. At moderate concurrency, the system remains memory I/O-bound, but since a sufficient number of tokens are processed simultaneously, reducing $n_a$ does not significantly reduce the total number of experts selected across all requests, resulting in limited acceleration.
At high concurrency, the system shifts to a compute-bound regime, where reducing $n_a$ lowers computational demands, thereby increasing throughput. In this case, the throughput gain aligns more closely with the compute reduction ratio.

\subsection{Performance and Structure Searching}

We aim to identify an inter-layer expert allocation strategy that maximizes model performance while maintaining a fixed total number of experts activated per token. In addition, we investigate the impact of different expert reduction strategies on model performance. Our current focus is on customized reductions at the layer level.

Specifically, we define expert allocation using a four-tuple $(b,h,e,p)$, where:
\begin{itemize}
    \item The first layer selects $b$ experts, i.e., $n_a(1)=b$.
    \item The $p$-th layer selects $h$ experts, i.e., $n_a(p)=h$.
    \item The final layer selects $e$ experts, i.e., $n_a(-1)=e$.
    \item For the other layers, expert counts are determined through linear interpolation.
\end{itemize}
This formulation allows us to explore various expert allocation patterns, including ascending, descending, peak, and valley-shaped distributions, depicted in Figure~\ref{fig:shapes}, Appendix~\ref{app:shapes}.
These experiments help us analyze the relative importance of experts across different layers from a language modeling perspective.

\begin{figure}[ht]
    \centering
    \begin{subfigure}{0.48\textwidth}
        \centering
        \caption{Deepseek-V2-Lite}
        \label{fig:v2lavgp}
        \includegraphics[width=\linewidth]{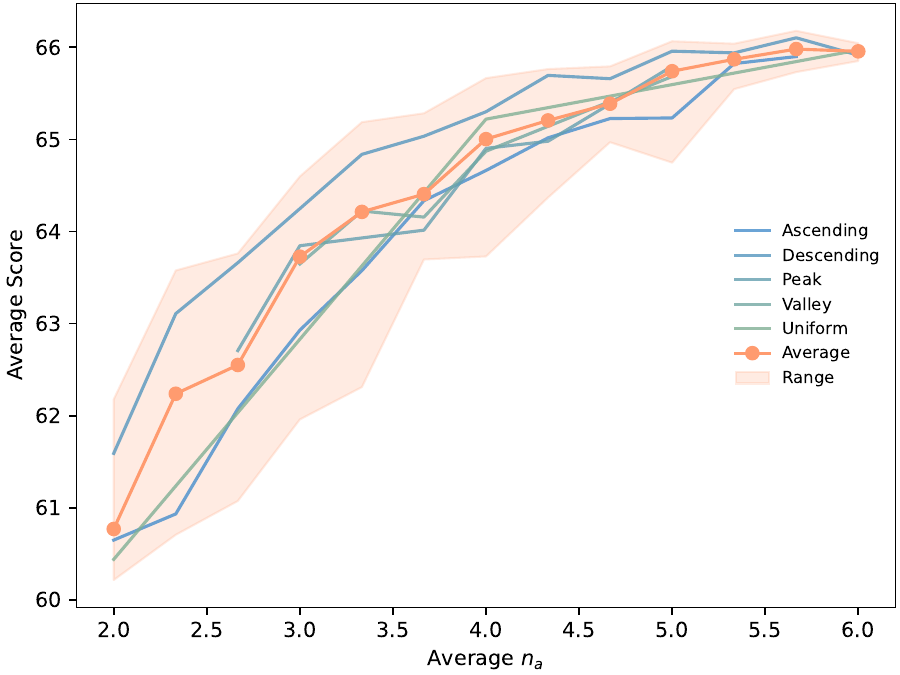}    
    \end{subfigure}
    \hfill
    \begin{subfigure}{0.48\textwidth}
        \centering
        \caption{Deepseek-V3}
        \label{fig:v3avgp}
        \includegraphics[width=\linewidth]{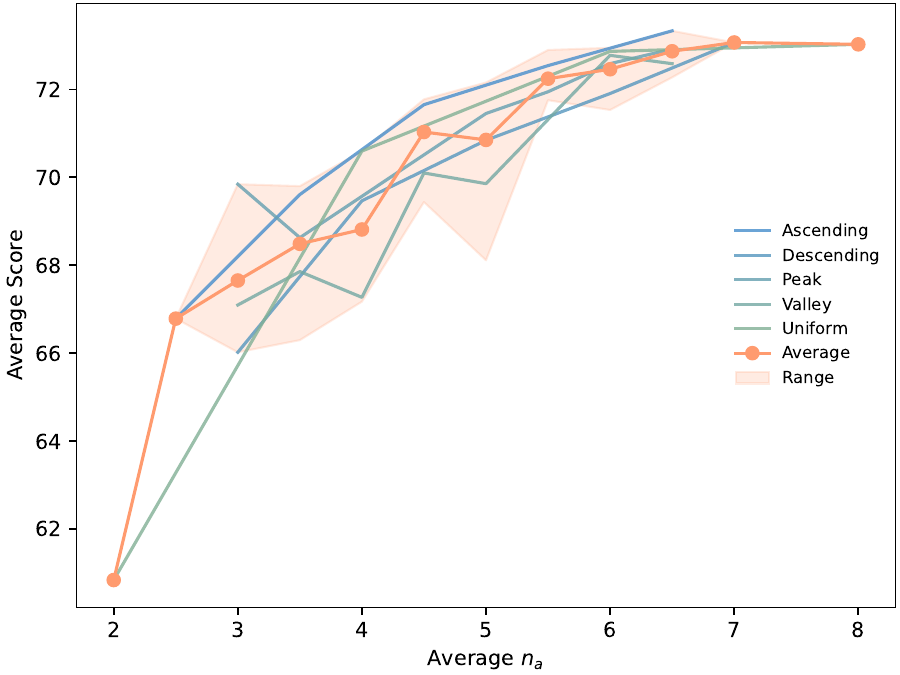}
    \end{subfigure}    
    \caption{How expert skipping influence performance}
    \label{fig:perftopk}
\end{figure}

Figure~\ref{fig:perftopk} presents our results. For softmax-based models such as DeepSeek V2, as shown in Figure~\ref{fig:v2lavgp}, we observe that even relatively aggressive expert skipping does not lead to a significant degradation in performance. On average, reducing the number of active experts from the full set to only two results in a performance drop of approximately 7.5\%, while in the best-case scenario, the performance loss is limited to around 6\%. Moreover, when an average of 3.3 experts is retained, the performance degradation remains within 1\%.


In the V3 model, as shown in Figure~\ref{fig:v3avgp}, we observe a similar performance curve when reducing $n_a$. However, compared to the smoother performance of V2-Lite, the V3 model exhibits greater instability across different reduction strategies. This instability may stem from the inherent properties of the sigmoid function, where expert weights tend to polarize toward 0 or 1. In contrast, in softmax-based models, the weights of lower-ranked experts are significantly smaller than that of the top-ranked expert. We argue that using sigmoid instead of softmax offers a greater advantage in fine-grained MoE models, as it enables more effective utilization of the selected experts. However, this also limits its applicability in expert skipping, if skipping removes experts whose weights have already converged to 1, a substantial performance drop may occur.

Furthermore, among various expert skipping strategies, our experiments indicate that the descending reduction strategy offers the best trade-off between efficiency and performance on V2-Lite, but the ascending strategy yields the best performance in V3.
We believe this behavior difference is an intrinsic characteristic of the model. The opposite trends observed in V2-Lite and V3 suggest that the optimal expert skipping strategy could be model-dependent, implying that a universal skipping strategy may not exist.

\section{Pre-Inference Expert Pruning}
\label{sec:6}

Beyond reducing the number of activated experts during inference, we also explore another possibility introduced by fine-grained MoE models, determining which experts to discard before inference begins. In other words, this corresponds to reducing the total number of experts ($n_e$). This approach has already been partially investigated, primarily in the context of Mixtral-series models, where expert pruning is typically guided by information-based metrics or performance comparisons after expert removal. 

However, we focus on two additional key questions: 1. To what extent can these methods accelerate inference? 2. Are these methods still effective for fine-grained MoE models? While reducing the total number of experts does not decrease the computational workload, it can increase computational intensity, which could theoretically lead to a net positive effect on inference speed. Furthermore, fine-grained MoE models introduce a unique challenge for global expert reduction, fully randomized initialization, meaning that experts do not exhibit any similarities, making effective global pruning significantly more complex.

\subsection{Efficiency}

\begin{figure}
    \centering
    \includegraphics[width=0.75\linewidth]{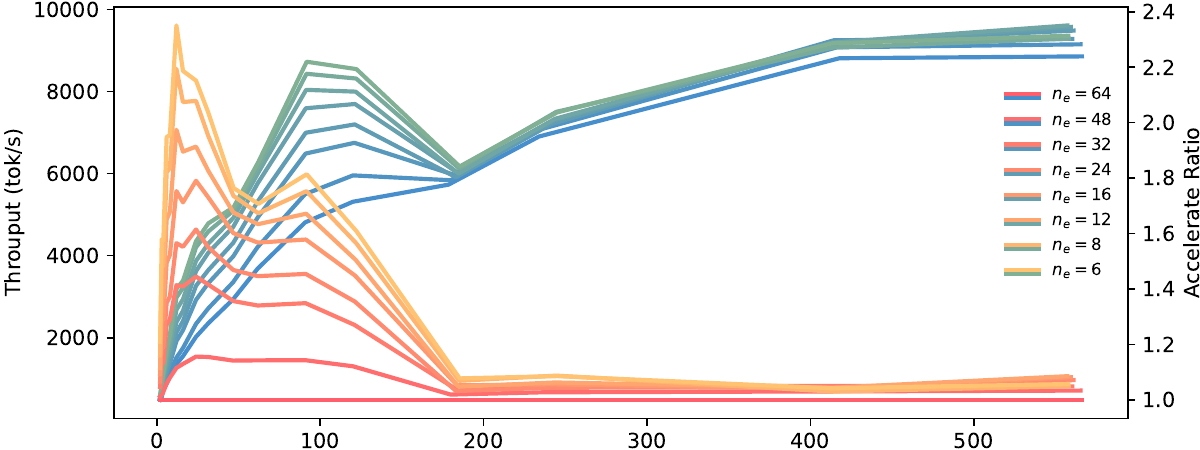}
    \caption{Throughput and speedup of expert pruning.}
    \label{fig:pmtpi}
\end{figure}

We conducted experiments on DeepSeek V2 with varying $n_e$ under a concurrent request setting of 512. 
The results of our evaluation are presented in Table~\ref{fig:pmtpi}.

We observed that the acceleration ratio was also significant at lower throughput levels, with up to 2.3× speedup. This is because when the number of experts is reduced, the computational intensity per expert increases rapidly, leading to a substantial increase in inference speed even with unchanged FLOPs. However, we noticed that at a concurrency level of 192, the throughput after expert reduction decreased. This may be due to a strategy shift within sglang for optimization, potentially triggering an underlying bug. For details on the sglang version used, please refer to Appendix~\ref{app:sec4}.

\subsection{Performance}

We evaluated the performance of different selection strategies under varying $n_e$ values, as presented in Table~\ref{tab:pmperf}. We conducted experiments using various expert reduction strategies, including random removal, structured removal (selecting experts based on odd indices, even indices, lower half indices, and upper half indices, methods theoretically equivalent to random selection), as well as the soft count and hard count methods proposed by \citeauthor{DBLP:journals/corr/abs-2404-05089}, which determine expert importance by there popularity before selecting the most critical ones. In some cases, certain methods completely lost their capabilities on specific test sets, performing at a level comparable to return random answer, we mark such cases in {\it italics}.

\begin{table}[htbp]
    \centering
    \begin{tabular}{l|c|cccccc|c}
    \toprule
        Method & 
         $n_e$ & ARC-C & ARC-E & BoolQ & OBQA & RTE & WinoGrande & Avg \\
    \midrule
        Baseline &
            64 & 52.9 & 80.6 & 83.1 & 35.8 & 73.3 & 71.4 & 66.0  \\
    \midrule
        \multirow{3}{1.5cm}{Random} &
            16 & {\it 20.8} & 27.2 & 60.8 & {\it 14.8} & {\it 50.5} & {\it 50.5} & {\it 37.4} \\
         &  32 & {\it 19.2} & 31.1 & 59.4 & {\it 15.2} & {\it 50.5} & {\it 51.2} & {\it 37.7} \\
         &  48 & 43.7 & 75.0 & 81.2 & 31.2 & 65.7 & 67.5 & 60.7 \\
    \midrule
        Odd &
            32 & {\it 22.2} & 30.9 & 65.9 & {\it 19.8} & 57.0 & 54.7 & 41.7 \\
        Even &
            32 & 32.1 & 62.2 & 69.4 & {\it 23.4} & 66.1 & 63.7 & 52.8 \\
        First half & 
            32 & {\it 21.4} & 30.8 & 62.5 & {\it 21.6} & 59.9 & {\it 52.7} & 41.5 \\
        Last half &
            32 & 32.6 & 60.1 & 75.3 & {\it 23.6} & 71.5 & 65.4 & 54.7 \\
    \midrule
        \multirow{3}{1.5cm}{Activate Count} &
            16 & {\it 21.8} & 31.4 & 62.0 & {\it 14.8} & 57.0 & {\it 50.7} & 39.6 \\
         &  32 & 40.8 & 68.3 & 75.0 & 29.2 & 54.9 & 70.2 & 56.3 \\
         &  48 & 44.7 & 75.8 & 79.4 & 33.8 & 66.4 & 72.6 & 62.1 \\
    \midrule
        \multirow{3}{1.5cm}{Soft Count} &
            16 & 28.7 & 57.3 & 62.2 & {\it 22.4} & 55.6 & 60.9 & 47.8 \\
         &  32 & 39.7 & 71.2 & 76.1 & 31.4 & 58.5 & 70.2 & 57.8 \\
         &  48 & 46.8 & 76.3 & 80.0 & 33.8 & 76.5 & 72.1 & 64.2 \\
    \bottomrule
    \end{tabular}
    \caption{Performance of different expert pruning method.}
    \label{tab:pmperf}
\end{table}

Our results indicate that the soft count method consistently outperformed the others. When reducing the number of experts by 25\%, even the best method exhibited some performance degradation, whereas random selection resulted in significant accuracy loss. With a 50\% reduction in experts, even the best method experienced a 15\% drop in performance, while randomly selected experts completely lost language capabilities. Finally, when only 25\% of the experts were retained, random selection still resulted in total failure, and only the best method was able to maintain some capability, while even the slightly inferior hard count method almost entirely lost its effectiveness.

An interesting observation from our experiments is that structured removal outperformed purely random selection. More specifically, when retaining experts with even indices or those from the latter half, performance was close to that of the best method. In contrast, the other two structured methods led to near-total language capability loss. This suggests the existence of particularly critical experts, which happen to be those with even indices and higher-numbered designations. In other words, despite the load-balancing mechanisms applied during training, there remains a significant disparity in expert importance.


\section{Conclusion}

Through our research and experiments, we have derived several key conclusions regarding fine-grained MoE models.

Compared to typical FFNs, MoE layers, despite having the same computational requirements, are more challenging to execute efficiently due to increased scheduling overhead and weaker batch-processing effects. However, expert parallelism offers potential for optimization.
Expert skipping during inference improves throughput. Although the presence of a shared backbone and attention mechanism constrains acceleration gains, small-batch and large-batch scenarios still exhibit significant improvements, whereas the effects remain less pronounced at intermediate concurrency levels. Encouragingly, the performance impact of expert skipping is minimal, and with appropriate skipping strategies, performance degradation can be further mitigated. Our best approach can increase throughput by at least 10\% without any loss in performance on Deepseek-V3.
On a global level, reducing the number of total experts before inference can yield a moderate increase in throughput, though the acceleration effect diminishes when the expert count is minimized. While reducing memory consumption lowers the deployment barrier, it also results in substantial performance loss, which limits its practical usability.

Overall, we believe that MoE optimization remains a promising research direction, both in terms of designing more efficient inference systems and exploring its potential from a language modeling perspective.

\bibliography{colm2025_conference}
\bibliographystyle{colm2025_conference}

\appendix
\section{Importancy of Controlling Input and Output Token Count}
\label{app:iosum}
We conducted tests under a concurrent request setting of 512, ensuring that the total number of input and output tokens remained fixed at 2048. The results of our evaluation are presented in Table~\ref{tab:iotp} for Deepseek-V2-Lite. Some datapoints on Deepseek V3 is presented in Table~\ref{tab:iotpv3}.

From this experiment, we observe that increasing the proportion of input tokens can significantly enhance overall throughput. This is primarily due to the fact that the prefill phase is highly parallelizable, with the computational cost evenly distributed across all tokens. In contrast, autoregressive decoding operates sequentially, processing only one token at a time, which leads to lower efficiency.

For the same reason, all our efficiency evaluations are conducted under this fixed setting rather than relying on performance benchmarks based on real-world workload scenarios. The latter approach introduces uncontrollable variables and lacks accuracy in assessing computational efficiency.

\begin{table}[htbp]
    \centering
    \begin{tabular}{l|ccccccc}
    \toprule
        Input token  &  256 &  512 &  768 & 1024 & 1280 & 1536 & 1792 \\
        Output token & 1792 & 1536 & 1280 & 1024 &  768 &  512 &  256 \\
    \midrule
        Throughput   & 6368 & 7106 & 8224 & 9484 & 10419& 11584& 13040 \\
    \bottomrule
    \end{tabular}
    \caption{Influence of IO token counts on throughput for Deepseek-V2-Lite.}
    \label{tab:iotp}
\end{table}

\begin{table}[htbp]
    \centering
    \begin{tabular}{l|ccccccc}
    \toprule
        Input token  & 1024 & 1024 \\
        Output token & 1024 &    8 \\
    \midrule
        Throughput   & 2636 & 8487 \\
    \bottomrule
    \end{tabular}
    \caption{Influence of IO token counts on throughput for Deepseek-V3.}
    \label{tab:iotpv3}
\end{table}

\section{Implementation Detail of Section~\ref{sec:3}}
\label{app:sec3}
\subsection{Hardware}

Please reference Table~\ref{tab:hdw}.

\begin{table}[ht]
    \centering
    \begin{tabular}{c|l|l}
        \toprule
        Item & Type & Quantity \\
        \midrule
        CPU & Intel Xeon Silver 4314 CPU @ 2.40GHz & 24 \\
        GPU & NVIDIA Tesla A800 80G PCI-e & 1 \\
        Memory & 16GB ECC DDR4@2666MHz & 15 \\
        \bottomrule
    \end{tabular}
    \caption{Hardware Information}
    \label{tab:hdw}
\end{table}

\subsection{Software}

We utilize pyTorch for the basic framework with {\tt torch.compile}. We use an implementation of SwiGLU for GLU, and the implementation from Transformers~\citep{wolf-etal-2020-transformers} {\tt MixtralModel} for MoE methods.

\section{Implementation Detail of Section~\ref{sec:4}}
\label{app:sec4}
\subsection{Deepseek-V2-Lite}

\subsubsection{Hardware}
Please reference Table~\ref{tab:hdw2}.

\begin{table}[ht]
    \centering
    \begin{tabular}{c|l|l}
        \toprule
        Item & Type & Quantity \\
        \midrule
        CPU & Intel Xeon Silver 4314 CPU @ 2.40GHz & 24 \\
        GPU & NVIDIA Tesla A800 80G PCI-e & 2 \\
        Memory & 16GB ECC DDR4@2666MHz & 15 \\
        \bottomrule
    \end{tabular}
    \caption{Hardware Information}
    \label{tab:hdw2}
\end{table}

\subsubsection{Software}

We utilize sglang build v0.4.4 post 1 (commit {\tt ad4e58bf67ec833ff4d036af5129ec6e1633efc4})  as the backend and sglang.bench for profiling.

\subsection{Deepseek-V3}

\subsubsection{Hardware}
Please reference Table~\ref{tab:hdw3}.

\begin{table}[ht]
    \centering
    \begin{tabular}{c|l|l}
        \toprule
        Item & Type & Quantity \\
        \midrule
        CPU & Intel Xeon Platinum 8558 CPU @ 2.10GHz & 48x2 \\
        GPU & NVIDIA Tesla H200 141G SXM5 & 8 \\
        Memory & 64GB ECC DDR4@2666MHz & 32 \\
        \bottomrule
    \end{tabular}
    \caption{Hardware Information}
    \label{tab:hdw3}
\end{table}

\subsubsection{Software}

We utilize sglang build v0.4.4 post 1 (commit {\tt ad4e58bf67ec833ff4d036af5129ec6e1633efc4}) as the backend and sglang.bench for profiling.

\section{Structure Shapes in Section~\ref{sec:4}}
\label{app:shapes}
\begin{figure}
    \centering
    \includegraphics[width=1\linewidth]{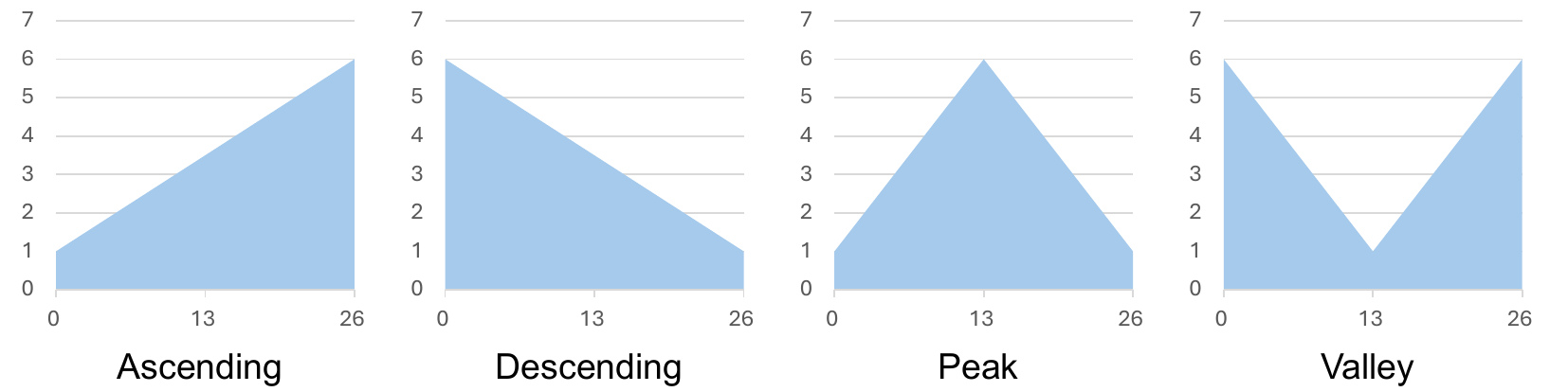}
    \caption{Structure shapes in Section~\ref{sec:4}.}
    \label{fig:shapes}
\end{figure}

Please reference Figure~\ref{fig:shapes}.

\section{Detailed Results of Section~\ref{sec:4} and \ref{sec:6}}
\label{app:dr1}

Reference Table~\ref{tab:fig3a} for Figure~\ref{fig:v2ltopk}.
Reference Table~\ref{tab:fig3b} for Figure~\ref{fig:v3topk}.
Reference Table~\ref{tab:fig4a} for Figure~\ref{fig:v2lavgp}.
Reference Table~\ref{tab:fig4b} for Figure~\ref{fig:v3avgp}.
Reference Table~\ref{tab:fig5} for Figure~\ref{fig:pmtpi}.

\begin{table}[htbp]
    \centering
    \begin{tabular}{l|ccccc}
    \toprule
         \multirow{2}{2.3cm}{Concurrency} & \multicolumn{5}{l}{$n_a=$} \\
         & 6 & 5 & 4 & 3 & 2 \\
    \midrule
        2 & 479 & 511 & 544 & 583 & 631 \\
        4 & 729 & 774 & 838 & 974 & 1099 \\
        8 & 1041 & 1122 & 1234 & 1380 & 1581 \\
        16 & 1519 & 1608 & 1738 & 1932 & 2254 \\
        32 & 2345 & 2412 & 2529 & 2716 & 3069 \\
        48 & 2968 & 3017 & 3111 & 3258 & 3572 \\
        64 & 3755 & 3801 & 3896 & 4042 & 4357 \\
        96 & 5020 & 5087 & 5172 & 5279 & 5608 \\
        128 & 5591 & 5660 & 5812 & 5960 & 6126 \\
        192 & 5948 & 6104 & 6069 & 6227 & 6461 \\
        256 & 7263 & 7385 & 7430 & 7501 & 7846 \\
        384 & 9052 & 9218 & 9369 & 9551 & 10137 \\
        512 & 9379 & 9783 & 9950 & 10102 & 10954 \\
        768 & 9453 & 9694 & 10043 & 10249 & 10968 \\
    \bottomrule
    \end{tabular}
    \caption{Detailed result of Figure~\ref{fig:v2ltopk}.}
    \label{tab:fig3a}
\end{table}

\begin{table}[htbp]
    \centering
    \begin{tabular}{l|cccc}
    \toprule
         \multirow{2}{2.3cm}{Concurrency} & \multicolumn{4}{l}{$n_a=$} \\
         & 8 & 6 & 4 & 3 \\
    \midrule
        2 & 163 & 164 & 167 & 170 \\
        4 & 300 & 308 & 312 & 323 \\
        8 & 526 & 537 & 554 & 575 \\
        16 & 769 & 793 & 820 & 979 \\
        32 & 1204 & 1331 & 1401 & 1705 \\
        64 & 1666 & 1751 & 1851 & 1900 \\
        128 & 2751 & 2806 & 2985 & 3100 \\
    \bottomrule
    \end{tabular}
    \caption{Detailed result of Figure~\ref{fig:v3topk}.}
    \label{tab:fig3b}
\end{table}

\begin{table}[htbp]
    \centering
    \begin{tabular}{c|cccccc|c}
    \toprule
         $(b,h,e,p)$ & ARC-C & ARC-E & BoolQ & OBQA & RTE & WinoGrande & Avg \\
    \midrule
        (2, 2, 2, 1) & 43.1 & 73.7 & 80.3 & 30.0 & 69.7 & 65.8 & 60.4 \\
(2, 2, 2, 6) & 43.6 & 73.9 & 80.3 & 30.0 & 69.7 & 66.9 & 60.7 \\
(2, 2, 2, 11) & 43.5 & 73.9 & 80.2 & 30.0 & 69.3 & 65.6 & 60.4 \\
(2, 2, 2, 16) & 43.5 & 74.0 & 80.3 & 30.0 & 69.7 & 65.0 & 60.4 \\
(2, 2, 2, 21) & 42.6 & 73.6 & 80.2 & 30.0 & 69.3 & 65.6 & 60.2 \\
(2, 2, 2, 26) & 43.3 & 73.4 & 80.2 & 30.4 & 69.7 & 65.5 & 60.4 \\
(2, 2, 4, 1) & 45.1 & 76.4 & 80.6 & 33.0 & 72.6 & 68.2 & 62.6 \\
(2, 2, 4, 6) & 45.1 & 76.0 & 79.7 & 31.0 & 71.8 & 66.3 & 61.6 \\
(2, 2, 4, 11) & 43.7 & 75.4 & 79.9 & 31.4 & 70.8 & 65.8 & 61.2 \\
(2, 2, 4, 16) & 43.5 & 74.8 & 80.2 & 29.6 & 70.8 & 65.4 & 60.7 \\
(2, 2, 4, 21) & 42.2 & 74.7 & 80.2 & 30.2 & 70.0 & 66.4 & 60.6 \\
(2, 2, 4, 26) & 42.8 & 73.5 & 80.5 & 30.2 & 69.7 & 64.6 & 60.2 \\
(2, 2, 6, 1) & 47.4 & 77.8 & 80.9 & 34.0 & 76.2 & 68.0 & 64.0 \\
(2, 2, 6, 6) & 46.5 & 77.0 & 80.3 & 32.2 & 70.8 & 67.1 & 62.3 \\
(2, 2, 6, 11) & 44.7 & 76.3 & 80.3 & 32.4 & 72.2 & 65.9 & 62.0 \\
(2, 2, 6, 16) & 43.4 & 75.0 & 80.0 & 30.6 & 71.1 & 66.2 & 61.1 \\
(2, 2, 6, 21) & 43.0 & 74.7 & 80.1 & 31.0 & 72.2 & 65.4 & 61.1 \\
(2, 2, 6, 26) & 43.5 & 73.8 & 80.2 & 29.8 & 71.1 & 65.7 & 60.7 \\
(2, 4, 2, 1) & 49.4 & 77.9 & 82.4 & 33.2 & 72.2 & 69.6 & 64.1 \\
(2, 4, 2, 6) & 48.3 & 77.7 & 82.9 & 32.8 & 71.8 & 69.9 & 63.9 \\
(2, 4, 2, 11) & 47.7 & 78.2 & 82.2 & 32.4 & 72.9 & 70.1 & 63.9 \\
(2, 4, 2, 16) & 48.2 & 77.2 & 81.5 & 33.6 & 73.3 & 68.7 & 63.8 \\
(2, 4, 2, 21) & 46.6 & 77.5 & 81.9 & 33.2 & 73.3 & 68.5 & 63.5 \\
(2, 4, 2, 26) & 46.4 & 76.9 & 80.7 & 33.0 & 72.6 & 66.7 & 62.7 \\
(2, 4, 4, 1) & 49.6 & 78.2 & 83.1 & 33.4 & 74.0 & 70.7 & 64.8 \\
(2, 4, 4, 6) & 48.0 & 78.5 & 82.4 & 33.4 & 74.7 & 70.5 & 64.6 \\
(2, 4, 4, 11) & 48.5 & 78.6 & 81.8 & 32.4 & 74.0 & 70.2 & 64.2 \\
(2, 4, 4, 16) & 47.6 & 77.5 & 81.2 & 34.8 & 75.1 & 68.8 & 64.2 \\
(2, 4, 4, 21) & 46.8 & 77.9 & 81.4 & 34.0 & 73.3 & 68.6 & 63.7 \\
(2, 4, 4, 26) & 45.6 & 76.6 & 80.7 & 33.4 & 71.1 & 67.3 & 62.4 \\
(2, 4, 6, 1) & 50.5 & 78.9 & 82.2 & 34.4 & 74.0 & 69.9 & 65.0 \\
(2, 4, 6, 6) & 48.4 & 78.5 & 82.5 & 32.6 & 74.7 & 70.9 & 64.6 \\
(2, 4, 6, 11) & 48.1 & 78.7 & 81.4 & 33.0 & 73.6 & 69.4 & 64.0 \\
(2, 4, 6, 16) & 48.0 & 77.7 & 81.2 & 33.6 & 74.4 & 69.8 & 64.1 \\
(2, 4, 6, 21) & 46.2 & 77.3 & 81.2 & 33.6 & 72.9 & 67.9 & 63.2 \\
(2, 4, 6, 26) & 46.2 & 76.6 & 80.9 & 33.2 & 71.8 & 66.6 & 62.6 \\
(2, 6, 2, 1) & 50.1 & 79.1 & 83.5 & 32.8 & 71.8 & 71.1 & 64.8 \\
(2, 6, 2, 6) & 48.5 & 78.5 & 82.8 & 34.4 & 74.0 & 69.9 & 64.7 \\
(2, 6, 2, 11) & 50.1 & 78.5 & 82.6 & 33.8 & 73.3 & 70.7 & 64.8 \\
(2, 6, 2, 16) & 48.7 & 78.4 & 82.8 & 33.6 & 73.3 & 70.3 & 64.5 \\
(2, 6, 2, 21) & 48.3 & 78.9 & 82.1 & 33.8 & 74.7 & 71.0 & 64.8 \\
(2, 6, 2, 26) & 49.4 & 77.7 & 81.3 & 34.8 & 73.6 & 68.3 & 64.2 \\
(2, 6, 4, 1) & 50.5 & 79.6 & 83.1 & 33.8 & 72.9 & 71.7 & 65.3 \\
(2, 6, 4, 6) & 50.3 & 79.3 & 82.8 & 34.4 & 73.6 & 71.6 & 65.3 \\
(2, 6, 4, 11) & 49.2 & 78.4 & 81.9 & 33.2 & 72.6 & 70.9 & 64.4 \\
(2, 6, 4, 16) & 48.4 & 79.1 & 82.4 & 33.6 & 75.1 & 70.4 & 64.8 \\
(2, 6, 4, 21) & 49.0 & 78.5 & 82.0 & 33.2 & 74.7 & 70.6 & 64.7 \\
(2, 6, 4, 26) & 48.9 & 77.8 & 81.2 & 34.4 & 72.6 & 68.2 & 63.8 \\
(2, 6, 6, 1) & 50.9 & 80.1 & 83.0 & 35.6 & 72.9 & 71.9 & 65.7 \\
(2, 6, 6, 6) & 51.0 & 79.5 & 82.4 & 34.4 & 74.0 & 72.0 & 65.5 \\
(2, 6, 6, 11) & 49.7 & 78.7 & 82.2 & 33.4 & 72.9 & 71.5 & 64.7 \\
(2, 6, 6, 16) & 48.8 & 78.7 & 82.1 & 34.0 & 75.1 & 71.1 & 65.0 \\
(2, 6, 6, 21) & 47.9 & 78.7 & 81.9 & 33.2 & 74.0 & 70.6 & 64.4 \\
(2, 6, 6, 26) & 48.6 & 77.5 & 81.3 & 34.2 & 75.1 & 68.0 & 64.1 \\
    \bottomrule
    \end{tabular}
\end{table}

\begin{table}[htbp]
    \centering
    \begin{tabular}{c|cccccc|c}
    \toprule
         $(b,h,e,p)$ & ARC-C & ARC-E & BoolQ & OBQA & RTE & WinoGrande & Avg \\
    \midrule
(4, 2, 2, 1) & 44.1 & 74.5 & 80.7 & 30.6 & 70.4 & 67.2 & 61.3 \\
(4, 2, 2, 6) & 45.5 & 75.6 & 81.0 & 31.4 & 71.8 & 67.7 & 62.2 \\
(4, 2, 2, 11) & 47.4 & 76.5 & 81.9 & 31.4 & 72.6 & 68.2 & 63.0 \\
(4, 2, 2, 16) & 48.4 & 77.1 & 82.2 & 32.6 & 73.3 & 67.9 & 63.6 \\
(4, 2, 2, 21) & 49.7 & 77.2 & 82.6 & 32.6 & 72.6 & 68.0 & 63.8 \\
(4, 2, 2, 26) & 49.1 & 77.7 & 82.5 & 31.8 & 72.9 & 69.3 & 63.9 \\
(4, 2, 4, 1) & 47.5 & 77.9 & 81.9 & 32.0 & 74.7 & 69.6 & 63.9 \\
(4, 2, 4, 6) & 47.7 & 76.4 & 80.7 & 30.6 & 73.3 & 68.0 & 62.8 \\
(4, 2, 4, 11) & 47.9 & 77.5 & 82.1 & 31.6 & 74.0 & 69.1 & 63.7 \\
(4, 2, 4, 16) & 47.4 & 77.5 & 81.8 & 32.0 & 75.1 & 69.5 & 63.9 \\
(4, 2, 4, 21) & 49.7 & 77.9 & 82.3 & 32.6 & 74.0 & 68.3 & 64.1 \\
(4, 2, 4, 26) & 49.0 & 78.4 & 82.3 & 32.2 & 72.6 & 69.3 & 63.9 \\
(4, 2, 6, 1) & 48.4 & 77.3 & 81.3 & 34.0 & 71.8 & 70.0 & 63.8 \\
(4, 2, 6, 6) & 48.5 & 77.6 & 81.1 & 32.6 & 73.6 & 68.8 & 63.7 \\
(4, 2, 6, 11) & 48.4 & 78.4 & 82.0 & 31.8 & 74.0 & 69.1 & 63.9 \\
(4, 2, 6, 16) & 47.9 & 78.1 & 82.1 & 32.8 & 74.0 & 69.3 & 64.0 \\
(4, 2, 6, 21) & 48.9 & 77.9 & 82.6 & 32.2 & 73.6 & 69.6 & 64.1 \\
(4, 2, 6, 26) & 49.2 & 78.0 & 82.6 & 32.0 & 72.6 & 68.4 & 63.8 \\
(4, 4, 2, 1) & 48.6 & 77.5 & 82.8 & 34.2 & 70.8 & 69.5 & 63.9 \\
(4, 4, 2, 6) & 50.1 & 78.4 & 82.8 & 32.8 & 72.2 & 70.5 & 64.5 \\
(4, 4, 2, 11) & 49.7 & 78.5 & 82.8 & 34.2 & 72.2 & 70.6 & 64.6 \\
(4, 4, 2, 16) & 50.2 & 77.9 & 82.7 & 34.2 & 73.3 & 69.9 & 64.7 \\
(4, 4, 2, 21) & 50.9 & 78.2 & 82.8 & 34.4 & 72.9 & 70.6 & 65.0 \\
(4, 4, 2, 26) & 50.5 & 79.0 & 82.8 & 33.6 & 73.3 & 70.4 & 64.9 \\
(4, 4, 4, 1) & 50.5 & 79.1 & 82.7 & 34.8 & 73.3 & 70.7 & 65.2 \\
(4, 4, 4, 6) & 50.7 & 79.4 & 82.8 & 34.8 & 73.3 & 70.3 & 65.2 \\
(4, 4, 4, 11) & 51.0 & 79.0 & 82.7 & 34.4 & 73.6 & 70.1 & 65.1 \\
(4, 4, 4, 16) & 51.3 & 79.0 & 82.7 & 34.8 & 73.3 & 70.3 & 65.2 \\
(4, 4, 4, 21) & 50.3 & 78.9 & 82.6 & 34.8 & 74.0 & 70.8 & 65.2 \\
(4, 4, 4, 26) & 50.9 & 79.0 & 82.7 & 35.2 & 73.6 & 70.5 & 65.3 \\
(4, 4, 6, 1) & 50.2 & 79.5 & 83.1 & 33.8 & 74.4 & 71.1 & 65.4 \\
(4, 4, 6, 6) & 51.6 & 79.0 & 82.9 & 33.6 & 73.6 & 70.9 & 65.3 \\
(4, 4, 6, 11) & 51.4 & 79.0 & 82.8 & 34.6 & 74.4 & 70.2 & 65.4 \\
(4, 4, 6, 16) & 51.2 & 79.0 & 82.9 & 33.8 & 73.3 & 70.2 & 65.1 \\
(4, 4, 6, 21) & 50.8 & 79.0 & 82.6 & 34.0 & 74.0 & 70.6 & 65.2 \\
(4, 4, 6, 26) & 50.4 & 78.9 & 82.7 & 34.4 & 73.3 & 70.6 & 65.1 \\
(4, 6, 2, 1) & 51.7 & 79.5 & 83.5 & 34.6 & 72.9 & 70.2 & 65.4 \\
(4, 6, 2, 6) & 50.9 & 79.5 & 83.5 & 35.0 & 72.6 & 71.9 & 65.6 \\
(4, 6, 2, 11) & 52.1 & 79.5 & 83.3 & 33.6 & 73.6 & 70.7 & 65.5 \\
(4, 6, 2, 16) & 51.8 & 78.9 & 83.4 & 33.8 & 73.3 & 70.2 & 65.2 \\
(4, 6, 2, 21) & 51.8 & 79.5 & 83.2 & 35.0 & 74.0 & 71.3 & 65.8 \\
(4, 6, 2, 26) & 50.8 & 79.5 & 83.1 & 33.4 & 73.3 & 71.3 & 65.2 \\
(4, 6, 4, 1) & 52.1 & 80.1 & 83.5 & 35.0 & 72.6 & 71.8 & 65.9 \\
(4, 6, 4, 6) & 52.5 & 80.0 & 83.1 & 35.0 & 74.0 & 71.8 & 66.1 \\
(4, 6, 4, 11) & 52.2 & 80.4 & 83.2 & 33.8 & 74.0 & 71.7 & 65.9 \\
(4, 6, 4, 16) & 52.1 & 79.5 & 83.1 & 34.0 & 74.0 & 70.7 & 65.6 \\
(4, 6, 4, 21) & 51.5 & 79.4 & 83.3 & 34.4 & 73.6 & 71.3 & 65.6 \\
(4, 6, 4, 26) & 50.7 & 79.6 & 83.0 & 33.6 & 73.6 & 71.3 & 65.3 \\
(4, 6, 6, 1) & 52.7 & 80.2 & 83.0 & 34.6 & 73.3 & 71.7 & 65.9 \\
(4, 6, 6, 6) & 52.2 & 80.1 & 83.2 & 34.8 & 73.3 & 72.7 & 66.1 \\
(4, 6, 6, 11) & 52.9 & 79.9 & 83.0 & 34.2 & 73.6 & 72.5 & 66.0 \\
(4, 6, 6, 16) & 51.9 & 79.5 & 83.1 & 35.2 & 74.0 & 71.6 & 65.9 \\
(4, 6, 6, 21) & 51.1 & 79.7 & 83.2 & 34.6 & 74.0 & 71.7 & 65.7 \\
(4, 6, 6, 26) & 51.2 & 79.7 & 83.3 & 33.8 & 74.0 & 71.3 & 65.6 \\
    \bottomrule
    \end{tabular}
\end{table}

\begin{table}[htbp]
    \centering
    \begin{tabular}{c|cccccc|c}
    \toprule
         $(b,h,e,p)$ & ARC-C & ARC-E & BoolQ & OBQA & RTE & WinoGrande & Avg \\
    \midrule
(6, 2, 2, 1) & 44.8 & 75.1 & 80.7 & 31.4 & 70.0 & 66.0 & 61.3 \\
(6, 2, 2, 6) & 46.3 & 75.8 & 81.7 & 32.4 & 72.2 & 68.0 & 62.7 \\
(6, 2, 2, 11) & 47.5 & 77.2 & 82.4 & 32.8 & 72.6 & 68.8 & 63.6 \\
(6, 2, 2, 16) & 49.5 & 78.4 & 83.0 & 34.0 & 72.6 & 70.2 & 64.6 \\
(6, 2, 2, 21) & 50.7 & 78.8 & 83.9 & 33.8 & 72.6 & 69.3 & 64.8 \\
(6, 2, 2, 26) & 50.8 & 79.3 & 83.5 & 34.4 & 71.5 & 71.6 & 65.2 \\
(6, 2, 4, 1) & 47.8 & 77.2 & 80.6 & 32.2 & 70.8 & 67.9 & 62.7 \\
(6, 2, 4, 6) & 48.0 & 77.3 & 81.3 & 32.4 & 73.6 & 67.9 & 63.4 \\
(6, 2, 4, 11) & 48.3 & 78.2 & 82.6 & 32.6 & 73.3 & 69.3 & 64.1 \\
(6, 2, 4, 16) & 48.5 & 78.5 & 83.0 & 34.0 & 74.7 & 70.5 & 64.9 \\
(6, 2, 4, 21) & 50.5 & 79.1 & 83.8 & 33.6 & 72.6 & 70.2 & 65.0 \\
(6, 2, 4, 26) & 51.5 & 79.4 & 83.7 & 35.4 & 72.2 & 71.3 & 65.6 \\
(6, 2, 6, 1) & 48.9 & 77.9 & 81.2 & 33.4 & 72.2 & 68.8 & 63.7 \\
(6, 2, 6, 6) & 49.0 & 77.9 & 81.4 & 32.8 & 74.4 & 69.5 & 64.2 \\
(6, 2, 6, 11) & 50.0 & 78.9 & 82.4 & 32.8 & 73.6 & 70.4 & 64.7 \\
(6, 2, 6, 16) & 50.2 & 79.3 & 83.1 & 34.4 & 74.7 & 71.2 & 65.5 \\
(6, 2, 6, 21) & 50.9 & 79.4 & 83.6 & 34.0 & 73.3 & 69.5 & 65.1 \\
(6, 2, 6, 26) & 51.0 & 79.7 & 83.7 & 35.2 & 71.1 & 71.3 & 65.3 \\
(6, 4, 2, 1) & 49.5 & 77.8 & 82.8 & 33.8 & 72.6 & 69.8 & 64.4 \\
(6, 4, 2, 6) & 50.8 & 79.1 & 83.5 & 33.6 & 74.0 & 70.2 & 65.2 \\
(6, 4, 2, 11) & 51.0 & 79.2 & 83.8 & 34.8 & 72.2 & 70.6 & 65.3 \\
(6, 4, 2, 16) & 51.7 & 80.0 & 83.6 & 33.6 & 72.6 & 70.1 & 65.3 \\
(6, 4, 2, 21) & 52.2 & 80.1 & 83.4 & 35.4 & 72.9 & 70.4 & 65.7 \\
(6, 4, 2, 26) & 52.6 & 80.3 & 83.5 & 35.6 & 72.6 & 71.7 & 66.1 \\
(6, 4, 4, 1) & 51.2 & 79.6 & 83.3 & 34.6 & 74.7 & 70.2 & 65.6 \\
(6, 4, 4, 6) & 50.7 & 79.9 & 83.2 & 35.0 & 74.0 & 71.2 & 65.7 \\
(6, 4, 4, 11) & 51.9 & 80.1 & 83.7 & 34.6 & 73.3 & 71.0 & 65.8 \\
(6, 4, 4, 16) & 51.9 & 80.5 & 83.5 & 34.4 & 72.9 & 70.6 & 65.6 \\
(6, 4, 4, 21) & 52.2 & 80.1 & 83.4 & 35.0 & 72.6 & 71.3 & 65.8 \\
(6, 4, 4, 26) & 52.4 & 80.2 & 83.4 & 35.6 & 72.6 & 71.6 & 65.9 \\
(6, 4, 6, 1) & 51.6 & 79.3 & 83.3 & 34.4 & 74.4 & 70.8 & 65.6 \\
(6, 4, 6, 6) & 51.4 & 79.4 & 83.3 & 34.0 & 74.0 & 72.3 & 65.7 \\
(6, 4, 6, 11) & 52.4 & 79.7 & 83.7 & 33.8 & 72.9 & 70.2 & 65.4 \\
(6, 4, 6, 16) & 51.5 & 80.3 & 83.6 & 34.4 & 73.3 & 71.1 & 65.7 \\
(6, 4, 6, 21) & 51.6 & 80.2 & 83.3 & 35.0 & 72.2 & 71.2 & 65.6 \\
(6, 4, 6, 26) & 52.1 & 80.2 & 83.4 & 35.8 & 72.6 & 71.9 & 66.0 \\
(6, 6, 2, 1) & 50.3 & 80.1 & 83.6 & 34.4 & 71.8 & 70.8 & 65.2 \\
(6, 6, 2, 6) & 52.0 & 80.4 & 83.5 & 35.2 & 72.6 & 70.5 & 65.7 \\
(6, 6, 2, 11) & 51.5 & 80.2 & 83.5 & 35.0 & 71.8 & 71.0 & 65.5 \\
(6, 6, 2, 16) & 52.3 & 80.0 & 83.6 & 36.2 & 71.8 & 71.3 & 65.9 \\
(6, 6, 2, 21) & 52.6 & 80.3 & 83.3 & 36.4 & 72.2 & 71.3 & 66.0 \\
(6, 6, 2, 26) & 52.7 & 80.5 & 83.2 & 35.2 & 72.6 & 71.3 & 65.9 \\
(6, 6, 4, 1) & 52.0 & 80.4 & 83.6 & 35.4 & 72.6 & 71.4 & 65.9 \\
(6, 6, 4, 6) & 52.3 & 80.4 & 83.3 & 36.0 & 72.2 & 71.7 & 66.0 \\
(6, 6, 4, 11) & 52.6 & 80.6 & 83.2 & 35.6 & 72.2 & 70.9 & 65.8 \\
(6, 6, 4, 16) & 52.7 & 80.4 & 83.3 & 35.2 & 72.6 & 72.0 & 66.0 \\
(6, 6, 4, 21) & 53.2 & 80.3 & 83.1 & 36.0 & 72.6 & 71.9 & 66.2 \\
(6, 6, 4, 26) & 53.2 & 80.3 & 83.1 & 35.6 & 72.6 & 70.6 & 65.9 \\
(6, 6, 6, 1) & 53.3 & 80.3 & 83.3 & 34.8 & 72.6 & 70.9 & 65.9 \\
(6, 6, 6, 6) & 52.8 & 80.5 & 83.2 & 35.6 & 72.9 & 70.6 & 65.9 \\
(6, 6, 6, 11) & 52.8 & 80.1 & 83.1 & 35.8 & 72.9 & 71.4 & 66.0 \\
(6, 6, 6, 16) & 52.9 & 80.0 & 83.1 & 35.8 & 73.3 & 71.2 & 66.0 \\
(6, 6, 6, 21) & 52.9 & 80.6 & 83.0 & 35.8 & 72.6 & 71.0 & 66.0 \\
(6, 6, 6, 26) & 52.9 & 80.5 & 83.1 & 35.4 & 72.9 & 71.0 & 66.0 \\
    \bottomrule
    \end{tabular}
    \caption{Detailed result of Figure~\ref{fig:v2lavgp}.}
    \label{tab:fig4a}
\end{table}

\begin{table}[htbp]
    \centering
    \begin{tabular}{c|cccccc|c}
    \toprule
         $(b,h,e,p)$ & ARC-C & ARC-E & BoolQ & OBQA & RTE & WinoGrande & Avg \\
    \midrule
(2, 2, 2, 61) & 53.2 & 78.9 & 69.9 & 33.2 & 65.3 & 64.5 & 60.8 \\
(2, 4, 4, 61) & 58.1 & 84.4 & 78.6 & 36.4 & 69.7 & 73.6 & 66.8 \\
(2, 6, 6, 61) & 63.1 & 85.4 & 83.8 & 39.2 & 71.1 & 76.2 & 69.8 \\
(2, 8, 8, 61) & 65.4 & 87.0 & 86.9 & 37.4 & 73.6 & 78.2 & 71.4 \\
(4, 2, 2, 61) & 57.9 & 82.5 & 78.9 & 35.0 & 69.3 & 72.5 & 66.0 \\
(4, 4, 4, 61) & 64.1 & 85.5 & 85.6 & 38.4 & 71.8 & 78.2 & 70.6 \\
(4, 6, 6, 61) & 65.5 & 86.8 & 87.3 & 38.6 & 73.6 & 78.8 & 71.8 \\
(4, 8, 8, 61) & 65.3 & 87.6 & 88.0 & 38.0 & 74.0 & 80.2 & 72.2 \\
(6, 2, 2, 61) & 61.9 & 85.5 & 84.1 & 36.2 & 73.3 & 75.1 & 69.3 \\
(6, 4, 4, 61) & 64.2 & 87.1 & 87.4 & 37.8 & 73.6 & 78.4 & 71.4 \\
(6, 6, 6, 61) & 65.8 & 87.1 & 88.5 & 39.8 & 76.2 & 79.9 & 72.9 \\
(6, 8, 8, 61) & 66.0 & 87.9 & 88.6 & 39.2 & 77.6 & 80.6 & 73.3 \\
(8, 2, 2, 61) & 62.8 & 85.4 & 87.3 & 37.4 & 74.0 & 77.2 & 70.7 \\
(8, 4, 4, 61) & 65.2 & 87.3 & 88.4 & 38.0 & 75.8 & 79.0 & 72.3 \\
(8, 6, 6, 61) & 66.5 & 87.8 & 89.0 & 39.6 & 76.2 & 79.4 & 73.1 \\
(8, 8, 8, 61) & 65.5 & 88.0 & 89.2 & 39.6 & 75.5 & 80.4 & 73.0 \\
(2, 5, 2, 10) & 61.3 & 85.4 & 83.5 & 34.6 & 74.0 & 74.5 & 68.9 \\
(2, 5, 2, 30) & 62.8 & 85.5 & 84.7 & 37.4 & 73.3 & 75.4 & 69.8 \\
(2, 5, 2, 50) & 62.4 & 85.0 & 81.9 & 36.4 & 70.8 & 73.8 & 68.4 \\
(2, 5, 8, 10) & 66.0 & 87.8 & 87.9 & 39.4 & 77.3 & 79.0 & 72.9 \\
(2, 5, 8, 30) & 65.9 & 87.3 & 86.9 & 38.2 & 73.6 & 78.7 & 71.8 \\
(2, 5, 8, 50) & 62.0 & 85.7 & 83.5 & 37.6 & 71.1 & 76.6 & 69.4 \\
(2, 8, 2, 10) & 64.7 & 85.9 & 87.0 & 36.8 & 73.6 & 78.1 & 71.0 \\
(2, 8, 2, 30) & 66.0 & 87.0 & 87.5 & 38.0 & 76.5 & 77.9 & 72.2 \\
(2, 8, 2, 50) & 64.2 & 86.1 & 86.1 & 37.4 & 72.6 & 78.5 & 70.8 \\
(2, 8, 5, 10) & 65.5 & 87.5 & 89.2 & 38.6 & 75.8 & 79.6 & 72.7 \\
(2, 8, 5, 30) & 65.8 & 87.4 & 88.5 & 37.6 & 74.4 & 79.2 & 72.1 \\
(2, 8, 5, 50) & 64.5 & 87.0 & 87.3 & 37.8 & 74.4 & 79.6 & 71.8 \\
(5, 2, 5, 10) & 59.9 & 85.0 & 80.5 & 34.8 & 69.0 & 73.4 & 67.1 \\
(5, 2, 5, 30) & 58.4 & 83.6 & 79.7 & 34.6 & 69.3 & 72.2 & 66.3 \\
(5, 2, 5, 50) & 60.7 & 85.4 & 82.6 & 36.0 & 73.6 & 73.4 & 68.6 \\
(5, 2, 8, 10) & 62.6 & 86.4 & 84.9 & 38.6 & 67.1 & 76.9 & 69.4 \\
(5, 2, 8, 30) & 60.0 & 84.7 & 81.2 & 36.8 & 66.8 & 73.6 & 67.2 \\
(5, 2, 8, 50) & 60.3 & 85.7 & 83.3 & 35.4 & 72.6 & 75.1 & 68.7 \\
(5, 8, 2, 10) & 64.0 & 85.7 & 87.0 & 38.4 & 74.7 & 79.1 & 71.5 \\
(5, 8, 2, 30) & 64.4 & 86.7 & 88.5 & 37.0 & 75.5 & 78.4 & 71.8 \\
(5, 8, 2, 50) & 64.9 & 87.2 & 89.0 & 37.0 & 75.5 & 79.0 & 72.1 \\
(5, 8, 5, 10) & 65.9 & 87.5 & 89.1 & 38.8 & 77.6 & 80.0 & 73.2 \\
(5, 8, 5, 30) & 65.5 & 87.7 & 89.1 & 39.6 & 75.8 & 79.9 & 72.9 \\
(5, 8, 5, 50) & 65.4 & 87.6 & 89.1 & 38.4 & 75.1 & 80.5 & 72.7 \\
(8, 2, 5, 10) & 61.6 & 84.8 & 80.4 & 36.4 & 70.8 & 72.7 & 67.8 \\
(8, 2, 5, 30) & 58.4 & 84.4 & 81.9 & 36.2 & 72.2 & 71.0 & 67.4 \\
(8, 2, 5, 50) & 62.5 & 86.6 & 87.1 & 37.6 & 74.0 & 76.8 & 70.8 \\
(8, 2, 8, 10) & 63.1 & 86.7 & 85.1 & 38.2 & 72.9 & 75.8 & 70.3 \\
(8, 2, 8, 30) & 59.2 & 84.0 & 83.9 & 36.2 & 72.6 & 72.8 & 68.1 \\
(8, 2, 8, 50) & 62.3 & 86.2 & 87.3 & 38.6 & 74.7 & 77.7 & 71.1 \\
(8, 5, 2, 10) & 62.2 & 84.3 & 83.1 & 37.2 & 74.7 & 75.9 & 69.6 \\
(8, 5, 2, 30) & 62.2 & 85.6 & 87.3 & 36.8 & 74.4 & 76.2 & 70.4 \\
(8, 5, 2, 50) & 65.9 & 86.3 & 87.9 & 38.2 & 72.6 & 78.4 & 71.5 \\
(8, 5, 8, 10) & 65.2 & 88.0 & 88.5 & 38.0 & 76.5 & 80.4 & 72.8 \\
(8, 5, 8, 30) & 65.2 & 87.2 & 88.7 & 38.8 & 74.7 & 79.0 & 72.3 \\
(8, 5, 8, 50) & 66.0 & 87.1 & 88.6 & 39.2 & 75.1 & 81.5 & 72.9 \\
    \bottomrule
    \end{tabular}
    \caption{Detailed result of Figure~\ref{fig:v3avgp}.}
    \label{tab:fig4b}
\end{table}

\begin{table}[htbp]
    \centering
    \begin{tabular}{l|cccccccc}
    \toprule
         \multirow{2}{2.3cm}{Concurrency} & \multicolumn{8}{l}{$n_e=$} \\
         & 64 & 48 & 32 & 24 & 16 & 12 & 8 & 6 \\
    \midrule
        2 & 488 & 494 & 510 & 520 & 533 & 544 & 566 & 590 \\
        3 & 563 & 572 & 635 & 667 & 723 & 767 & 836 & 889 \\
        4 & 742 & 763 & 839 & 882 & 958 & 1020 & 1101 & 1169 \\
        6 & 876 & 928 & 1114 & 1182 & 1309 & 1446 & 1595 & 1708 \\
        8 & 1060 & 1144 & 1367 & 1455 & 1614 & 1779 & 1947 & 2072 \\
        12 & 1357 & 1513 & 1918 & 2124 & 2379 & 2680 & 2978 & 3190 \\
        16 & 1548 & 1749 & 2182 & 2403 & 2651 & 2935 & 3212 & 3386 \\
        24 & 2022 & 2338 & 2922 & 3265 & 3621 & 3870 & 4203 & 4351 \\
        32 & 2351 & 2714 & 3328 & 3645 & 4068 & 4292 & 4588 & 4786 \\
        48 & 2937 & 3352 & 3982 & 4312 & 4703 & 4920 & 5097 & 5180 \\
        64 & 3679 & 4203 & 4931 & 5321 & 5766 & 6011 & 6155 & 6280 \\
        96 & 4810 & 5497 & 6486 & 6993 & 7590 & 8038 & 8430 & 8723 \\
        128 & 5312 & 5954 & 6748 & 7196 & 7693 & 7994 & 8314 & 8543 \\
        192 & 5729 & 5836 & 5959 & 5909 & 6027 & 5999 & 6122 & 6171 \\
        256 & 6901 & 7087 & 7238 & 7202 & 7286 & 7333 & 7502 & 7490 \\
        512 & 8808 & 9074 & 9248 & 9201 & 9154 & 9191 & 9069 & 9173 \\
        784 & 8853 & 9147 & 9280 & 9481 & 9571 & 9607 & 9285 & 9356 \\
    \bottomrule
    \end{tabular}
    \caption{Detailed result of Figure~\ref{fig:pmtpi}.}
    \label{tab:fig5}
\end{table}

\end{document}